\documentclass[letterpaper, 10 pt, conference]{ieeeconf}  % Comment this line out if you need a4paper

\IEEEoverridecommandlockouts                              % This command is only needed if 
\usepackage{graphics} % for pdf, bitmapped graphics files
\usepackage{epsfig} % for postscript graphics files
\usepackage{mathptmx} % assumes new font selection scheme installed
\usepackage{times} % assumes new font selection scheme installed
\usepackage{amsmath} % assumes amsmath package installed
\usepackage{amssymb}  % assumes amsmath package installed
\usepackage{booktabs}
\usepackage{algorithm}
\usepackage{algorithmic}
\usepackage{multirow}
\usepackage{array}
\usepackage{xcolor}
\usepackage{cite}
\usepackage{float}
\usepackage{booktabs}
\usepackage{array}
\usepackage{tabularx}
\usepackage{siunitx} % 用于数字对齐
\usepackage{atbegshi}  % 用于在文档开始时执行命令
\usepackage{hyperref}

\AtBeginDocument{% 在文档开始时设置图计数器
  \setcounter{figure}{1}
}

\title{\LARGE \bf
CLEA: Closed-Loop Embodied Agent for Enhancing Task Execution in Dynamic Environments
}

\author{Mingcong Lei$^{*}$, Ge Wang$^{*}$, Yiming Zhao, Zhixin Mai, Qing Zhao, Yao Guo, \\Zhen Li, Shuguang Cui, Yatong Han$^{\dagger}$, and Jinke Ren$^{\dagger}$% <-this % stops a space
\thanks{* These authors contributed equally to this work.}% <-this % stops a space
\thanks{$\dagger$ Corresponding authors: Jinke Ren ({\tt{jinkeren@cuhk.edu.cn}}); Yatong Han ({\tt{hanyatong@cuhk.edu.cn}})}% <-this % stops a space
\thanks{M. Lei, G. Wang, Y. Zhao, Z. Mai, Q. Zhao, Y. Han are with the Shenzhen Future Network of Intelligence Institute (FNii-Shenzhen) and Guangdong Provincial Key Laboratory of Future Networks of Intelligence, The Chinese University of Hong Kong, Shenzhen 518172, China, and also with Infused Synapse AI, Shenzhen 518048, China. Y. Zhao is also with Harbin Engineering University, Harbin 15006, China.}
\thanks{Yao Guo is with the Institute of Medical Robotics, School of Biomedical Engineering, Shanghai Jiao Tong University, Shanghai 200240, China.}
\thanks{Z. Li, S. Cui and J. Ren are with the School of Science and Engineering (SSE), FNii-Shenzhen, and Guangdong Provincial Key Laboratory of Future Networks of Intelligence, The Chinese University of Hong Kong, Shenzhen 518172, China.}}

\begin{document}

\makeatletter
\let\@oldmaketitle\@maketitle
\renewcommand{\@maketitle}{\@oldmaketitle%
\vspace{4pt}
  \begin{center}
    \includegraphics[width=0.99\linewidth]{img/begining.jpg}    \label{fig:overview}
  \end{center}
  \vspace{-8pt}
  {\small
  Fig.~1: \textbf{Task execution processes of CLEA.} (1) Search task: CLEA facilitates adaptive strategy adjustments in partially observable environments. By analyzing visual inputs from robot 2, which remains stationary on the table, CLEA directs robot 1, the mobile unit, to explore alternative locations—such as the interior of the refrigerator—in search of water. (2) Manipulation task: CLEA evaluates the feasibility of each action and dynamically refines its strategy. Upon opening the oven, it determines that placing the apple directly inside is infeasible. Consequently, CLEA adjusts the next step to a more appropriate action: pulling out the grill to create sufficient space, thereby successfully completing the manipulation task. }
  \vspace{-8pt}
  \medskip}%
\makeatother

\maketitle

\thispagestyle{empty}
\pagestyle{empty}

%%%%%%%%%%%%%%%%%%%%%%%%%%%%%%%%%%%%%%%%%%%%%%%%%%%%%%%%%%%%%%%%%%%%%%%%%%%%%%%%

\begin{abstract}
Large Language Models (LLMs) exhibit remarkable capabilities in the hierarchical decomposition of complex tasks through semantic reasoning. However, their application in embodied systems faces challenges in ensuring reliable execution of subtask sequences and achieving one-shot success in long-term task completion. To address these limitations in dynamic environments, we propose Closed-Loop Embodied Agent (CLEA)---a novel architecture incorporating four specialized open-source LLMs with functional decoupling for closed-loop task management. The framework features two core innovations: (1) Interactive task planner that dynamically generates executable subtasks based on the environmental memory, and (2) Multimodal execution critic employing an evaluation framework to conduct a probabilistic assessment of action feasibility, triggering hierarchical re-planning mechanisms when environmental perturbations exceed preset thresholds. To validate CLEA's effectiveness, we conduct experiments in a real environment with manipulable objects, using two heterogeneous robots for object search, manipulation, and search-manipulation integration tasks. Across 12 task trials, CLEA outperforms the baseline model, achieving a 67.3\% improvement in success rate and a 52.8\% increase in task completion rate. These results demonstrate that CLEA significantly enhances the robustness of task planning and execution in dynamic environments. Our code is available at \href{https://sp4595.github.io/CLEA/}{https://sp4595.github.io/CLEA/}. 
\end{abstract}

%%%%%%%%%%%%%%%%%%%%%%%%%%%%%%%%%%%%%%%%%%%%%%%%%%%%%%%%%%%%%%%%%%%%%%%%%%%%%%%%
\section{Introduction}
%In recent years, Large Language Models (LLMs) have achieved significant advancements in natural language understanding and hierarchical task decomposition, demonstrating strong capabilities in high-dimensional embodied task planning \cite{zeng2022socratic}. LLMs can effectively interpret complex language instructions and generate executable action sequences, enabling multi-step tasks such as object rearrangement, home cleaning, and interactive cooking \cite{huang2022inner}. However, LLMs face fundamental challenges in handling long-horizon tasks due to context window limitations, which prevent them from continuously tracking and updating task states over extended sequences. Additionally, the static action sequences generated by LLMs lack adaptability to dynamic environments, where object states, obstacle configurations, and spatial relationships may change unpredictably. These limitations significantly hinder the robustness and adaptability of LLM-driven task planning in real-world applications \cite{jiang2022vima, ahn2022can}.

In recent years, Large Language Models (LLMs) have advanced significantly in natural language understanding and hierarchical task decomposition, demonstrating strong capabilities in embodied task planning \cite{zeng2022socratic}. They can interpret complex instructions and generate executable action sequences for multi-step tasks such as object rearrangement and interactive cooking \cite{huang2022inner}. However, LLMs struggle with long-horizon tasks due to context window limitations, preventing continuous task state tracking. Additionally, their static action sequences lack adaptability to dynamic environments where object states and spatial relationships change unpredictably, hindering robustness in real-world applications \cite{jiang2022vima, ahn2022can}.

%To address these challenges, recent studies have explored augmenting LLMs with external planning frameworks to enhance their performance in long-horizon tasks. One approach involves translating LLM-generated plans into structured representations such as Planning Domain Definition Language (PDDL) \cite{zhou2024isr} or Behavior Trees (BTs) \cite{ao2024llm}, improving task planning precision and structural consistency. However, these methods heavily rely on static task definitions and prior knowledge, limiting their adaptability to real-time environmental changes. Additionally, BTs-based task plans often require manual fine-tuning, further constraining their applicability in dynamic scenarios.

To overcome these challenges, recent work integrates LLMs with external planning frameworks. One approach translates LLM-generated plans into structured formats like Planning Domain Definition Language (PDDL) \cite{zhou2024isr} or Behavior Trees (BTs) \cite{ao2024llm}, enhancing precision but relying on static task definitions and manual tuning. Another direction combines LLMs with dynamic planning techniques, such as Monte Carlo Tree Search (MCTS) \cite{zhao2024large} for commonsense-driven reasoning, though its iterative search constraints limit scalability. Similarly, LLM-Task and Motion Planning (TAMP) \cite{yang2024guiding} improves task feasibility but faces exponential planning complexity in multi-agent systems, complicating coordination in dynamic environments.

%Another research direction integrates dynamic planning techniques with LLMs. For example, combining LLMs with Monte Carlo Tree Search (MCTS) \cite{zhao2024large} enables commonsense-driven task reasoning by guiding search processes. However, due to the limited number of strategy steps MCTS can return per iteration, this approach still requires extensive exploration, particularly in large-scale environments. Similarly, Task and Motion Planning (TAMP) algorithms, when combined with LLMs, improve task feasibility in complex home environments \cite{yang2024guiding}. However, in multi-agent systems, the global planning complexity of these layered methods scales exponentially with the number of agents, making it increasingly difficult to coordinate collaborative tasks in highly dynamic environments.

To address these limitations, we propose Closed-Loop Embodied Agents (CLEA), a novel framework designed to enable adaptive decision-making in dynamic, multi-robot environments. Its effectiveness across diverse tasks is demonstrated in Fig. 1. CLEA establishes a closed-loop planning paradigm by integrating four decoupled LLM modules, allowing for continuous adaptation based on real-time environmental feedback. The key contributions of this work are as follows.
\begin{itemize}
    \item \textbf{Closed-loop planner-critic reasoning:} We propose a novel planner-critic architecture that establishes a closed-loop perception-reasoning-execution framework. The planner generates temporally coherent action sequences by leveraging historical environmental states and summarized memory, ensuring consistency with real-world dynamics. Meanwhile, the critic conducts online feasibility assessments through event-triggered evaluations, enabling adaptive task execution based on robot capabilities and sub-task constraints. 
    \item \textbf{Open-source VLMs and LLMs powered agent framework:} To enhance reproducibility and extensibility, we develop the agent system using open-source Visual Language Models (VLMs) and LLMs. The entire framework is successfully deployed on real robotic platforms, demonstrating its practicality in real-world applications.  
    \item \textbf{Experiment evaluation in real-world environment:} We evaluate CLEA in a real-world setting, where two robots execute three types of complex, long-horizon tasks: object search, multi-object manipulation, and integration tasks. Experimental results indicate that CLEA significantly outperforms the open-loop baseline, achieving substantial improvements in task planning accuracy and execution robustness.

%We conduct open-source implementation. The entire system is developed using open-source Visual Language Models (VLMs) and LLMs, fostering reproducibility and extensibility within the research community.
\end{itemize}

\section{Related Work}
%\textbf{LLMs for Embodied Task Planning:} After the initial understanding of human intent, LLMs can effectively decompose high-dimensional embodied tasks. Current embodied task planning based on LLMs primarily employs two paradigms: open-loop planning and closed-loop planning. In open-loop systems (e.g., HuggingGPT)\cite{shen2024hugginggpt}, a complete action sequence is generated through a single inference, which offers advantages in task efficiency but struggles to adapt to changes in dynamic environments. Closed-loop systems (e.g., LLM-planner)\cite{song2023llm}, on the other hand, improve adaptability through iterative replanning, allowing adjustments based on environmental feedback during task execution, thereby enhancing task robustness. Both approaches have their respective strengths and weaknesses in terms of task efficiency and robustness. Previous studies have attempted to improve the feasibility of robot action sequences by incorporating multimodal prompt inputs. However, in the embodied domain, most existing research integrates other algorithms, such as PDDL, BTs, MCTS, and TAMP, using hierarchical planning frameworks to further enhance planning performance\cite{zhou2024isr}\cite{ao2024llm}\cite{zhao2024large}\cite{yang2024guiding}. While these methods improve planning accuracy and standardization to some extent, they typically rely on static task definitions and prior knowledge, lacking the ability to respond to dynamic environmental changes in real-time.
\textbf{LLMs for embodied task planning:} LLMs can effectively decompose complex embodied tasks after understanding human intent. Existing approaches follow two paradigms: open-loop and closed-loop planning. Open-loop methods (e.g., HuggingGPT \cite{shen2024hugginggpt}) generate complete action sequences in a single inference, offering efficiency but lacking adaptability in dynamic environments. Closed-loop systems (e.g., LLM-Planner \cite{song2023llm}) iteratively replan based on environmental feedback, improving robustness at the cost of efficiency. However, existing closed-loop systems have stringent requirements for the quality of environmental feedback. As a result, most current approaches can only achieve successful execution in simulated environments. To enhance feasibility, prior work integrates multimodal prompts and hierarchical frameworks with PDDL, BTs, MCTS, and TAMP \cite{zhou2024isr,ao2024llm,zhao2024large,yang2024guiding}. While these methods improve accuracy and standardization, they rely on static task definitions and prior knowledge, limiting real-time adaptability.

%\textbf{Environment Description with VLM:} In robotic systems, vision is typically a key sensory input used in conjunction with visual segmentation and object recognition techniques to define target objects. VLMs, with their multimodal capabilities, can effectively leverage language to describe the spatial relationships between items in a scene\cite{fu2024can}. By combining the textual output functionality of VLMs, researchers have begun to explore task planning in simulated environments by decoupling feedback from the environment engine, aiming to gradually reduce the gap between simulated and real-world scenarios\cite{wu2023embodied}\cite{yang2024guiding}. This approach offers new possibilities for autonomous task planning in complex environments, but its application in real-world scenarios still faces significant challenges. 

\textbf{Environment description with VLMs:} Vision is crucial in robotic systems, aiding object recognition and segmentation. VLMs enhance this by describing spatial relationships through language \cite{fu2024can}. Leveraging their textual output, researchers explore task planning in simulated environments by decoupling environmental feedback, narrowing the sim-to-real gap \cite{wu2023embodied, yang2024guiding}. While promising for autonomous planning in complex settings, real-world deployment remains challenging.

%\textbf{Multi-Agent Collaboration:} The core objective of building multi-agent systems is to solve complex tasks by appropriately distributing different responsibilities\cite{li2025large}\cite{singh2024twostep}. In such systems, agents may have similar or distinct capabilities. For instance, in large-scale navigation tasks, all robots might only possess mobility capabilities, while in inspection tasks, robots need to have the ability to quickly avoid obstacles and perform operational checks\cite{kannan2024smart}. In complex environments like kitchens, an important research direction in the field of embodied intelligence is how to collect the interactions of heterogeneous agents in dynamic and partially observable environments and autonomously update system strategies. 

\textbf{Multi-agent collaboration:} Multi-agent systems tackle complex tasks by distributing responsibilities among agents with similar or distinct capabilities \cite{li2025large, singh2024twostep}. In large-scale navigation, agents may share mobility functions, while inspection tasks require obstacle avoidance and operational checks \cite{kannan2024smart}. In dynamic, partially observable environments like kitchens, a key challenge in embodied intelligence is autonomously updating strategies based on heterogeneous agent interactions.

\section{Problem Formulation}
\label{sec:problem_formulation}

%In traditional robotic systems, errors during task execution are typically addressed through predefined correction strategies, manually designed to handle specific types of errors. However, this approach becomes increasingly inadequate when applied to high-dimensional, embodied tasks in partially observable environments, where it is unrealistic to anticipate all possible error scenarios in advance. In these contexts, robots must be capable of adapting to unforeseen challenges by dynamically adjusting their strategies during task execution.

Traditional robotic systems rely on predefined error correction strategies, which become inadequate for high-dimensional, embodied tasks in partially observable environments. Anticipating all possible errors in advance is impractical, requiring robots to dynamically adapt during execution.

To address this, we model the problem using a Partially Observable Markov Decision Process (POMDP) \cite{kaelbling1998planning}. Robots perceive the environment through onboard cameras, receiving partial sensory inputs as JPEG images. The true state remains hidden, introducing decision-making uncertainty. Robots interact via predefined action APIs, mapping to low-level operations. Formally, the POMDP is defined by the tuple \((S, A, \Omega, T, O)\), where \(S\) denotes the state space, \(A\) is the action space of executable function calls, \(\Omega\) is the visual observation space, \(T(s'|s, a)\) is the state transition dynamics, and \(O(o|s', a)\) is the observation function that generates camera images from environmental states. At each time step \(i\), the agent receives a visual observation \(o_i \sim O(\cdot|s_i, a_{i-1})\) and maintains a belief state \(b_i \in \Delta(S)\)---a probability distribution over latent states conditioned on the interaction history:

%To address this limitation, we model the problem using the Partially Observable Markov Decision Process (POMDP) \cite{kaelbling1998planning}. In this framework, robots perceive their environment through onboard cameras, which provide partial sensory information as JPEG images. The true state of the environment is hidden, introducing uncertainty into the robot's decision-making process. The robots interact with the environment via predefined action APIs, each corresponding to low-level operations on the robotic platform. These observations and actions are formalized in the POMDP framework, enabling robots to update their belief state and adjust strategies based on incomplete information. The POMDP is defined by the tuple \((S, A, \Omega, T, O)\), where \(S\) denotes the state space, \(A\) is the action space of executable function calls, \(\Omega\) is the visual observation space, \(T(s'|s, a)\) is the state transition dynamics, and \(O(o|s', a)\) is the observation function that generates camera images from environmental states, at each time step \(i\), the agent receives a visual observation \(o_i \sim O(\cdot|s_i, a_{i-1})\) and maintains a belief state \(b_i \in \Delta(S)\) — a probability distribution over latent states conditioned on the interaction history:

\begin{equation}
    b_i = P(s_i = s | o_{1:i}, a_{1:i-1}).
\end{equation}

%Following predictive processing principles \cite{kaelbling1998planning}, this belief state serves as the agent's internal estimation of the actual environmental state, continuously updated through sequential visual observations and action outcomes. The image-based observation space captures the embodiment constraints, as the robot's egocentric perspective inherently limits environmental perceptibility. This formulation enables principled reasoning under uncertainty while maintaining compatibility with real robotic systems' perceptual characteristics. Our goal is to generate \(b_i\)  based on the previous trajectory and then plan a series of actions within the closed loop. Next, using \(b_i\) , we will plan further actions to accomplish the task. 

Following the principles of predictive processing \cite{kaelbling1998planning}, the belief state serves as the agent's internal estimate of the true environmental state, continuously refined through sequential visual observations and action outcomes. The image-based observation space inherently reflects embodiment constraints, as the robot's egocentric perspective limits its perceptual field. This formulation provides a principled approach to reasoning under uncertainty while aligning with the perceptual characteristics of real-world robotic systems. Our objective is to generate 
\(b_i\) based on the prior trajectory and subsequently formulate a sequence of actions within the closed-loop framework. Using \(b_i\), the agent iteratively plans further actions to achieve the designated task.

\begin{figure}[t]
\centering

   \includegraphics[width=0.99\linewidth]{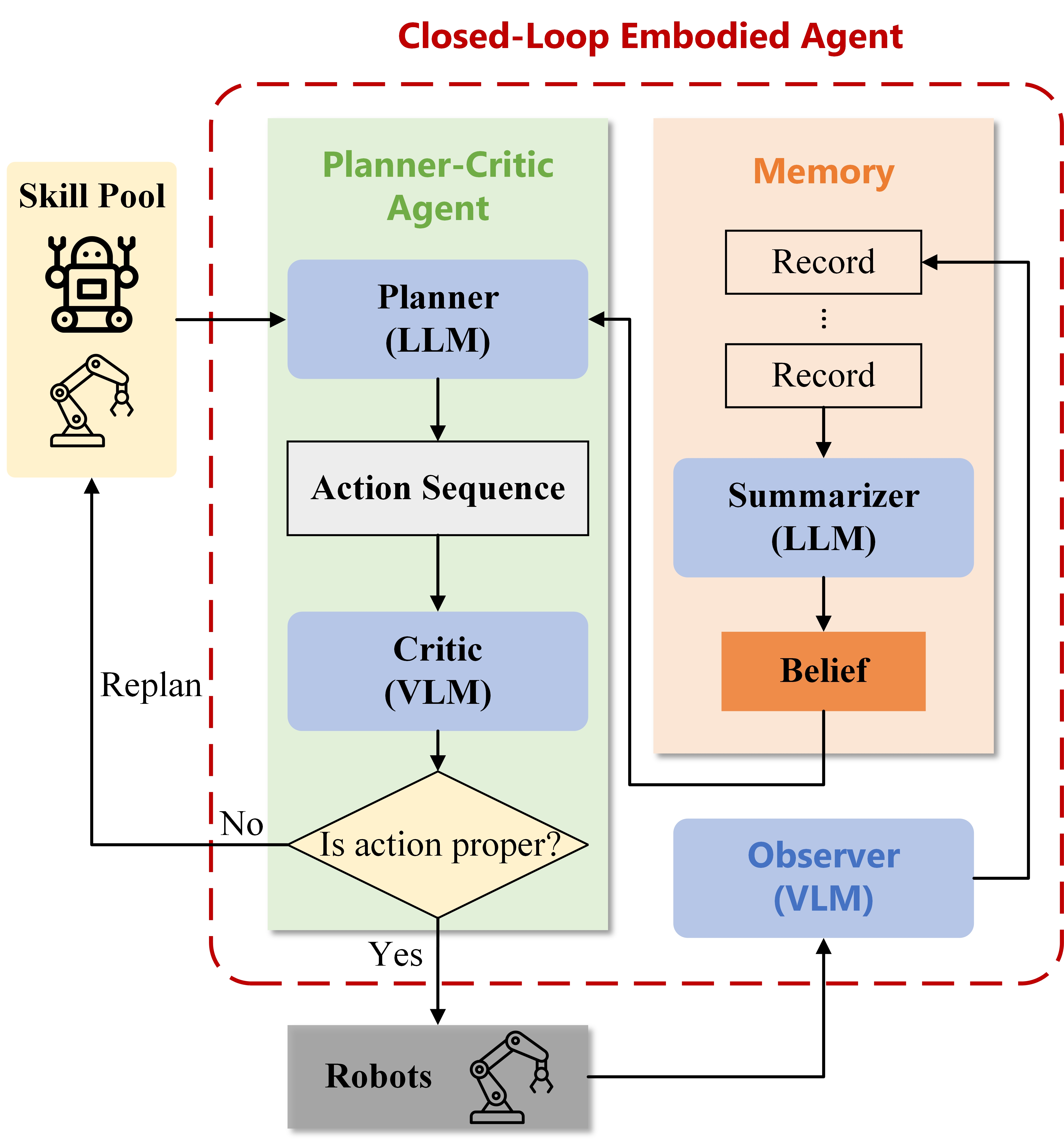}
   \vskip -0.05in
   \caption{\textbf{Overview of CLEA.} The observer (VLM) provides environmental data, which the summarizer (LLM) processes into memory. The planner (LLM) generates an initial action sequence based on the robot's skill pool and memory, while the critic (VLM) evaluates action feasibility and offers re-plan recommendations in response to environmental dynamics.}
   \label{fig:onecol}
\vskip -0.2in
\end{figure}
\begin{figure}[t]
\centering
   \includegraphics[width=0.99\linewidth]{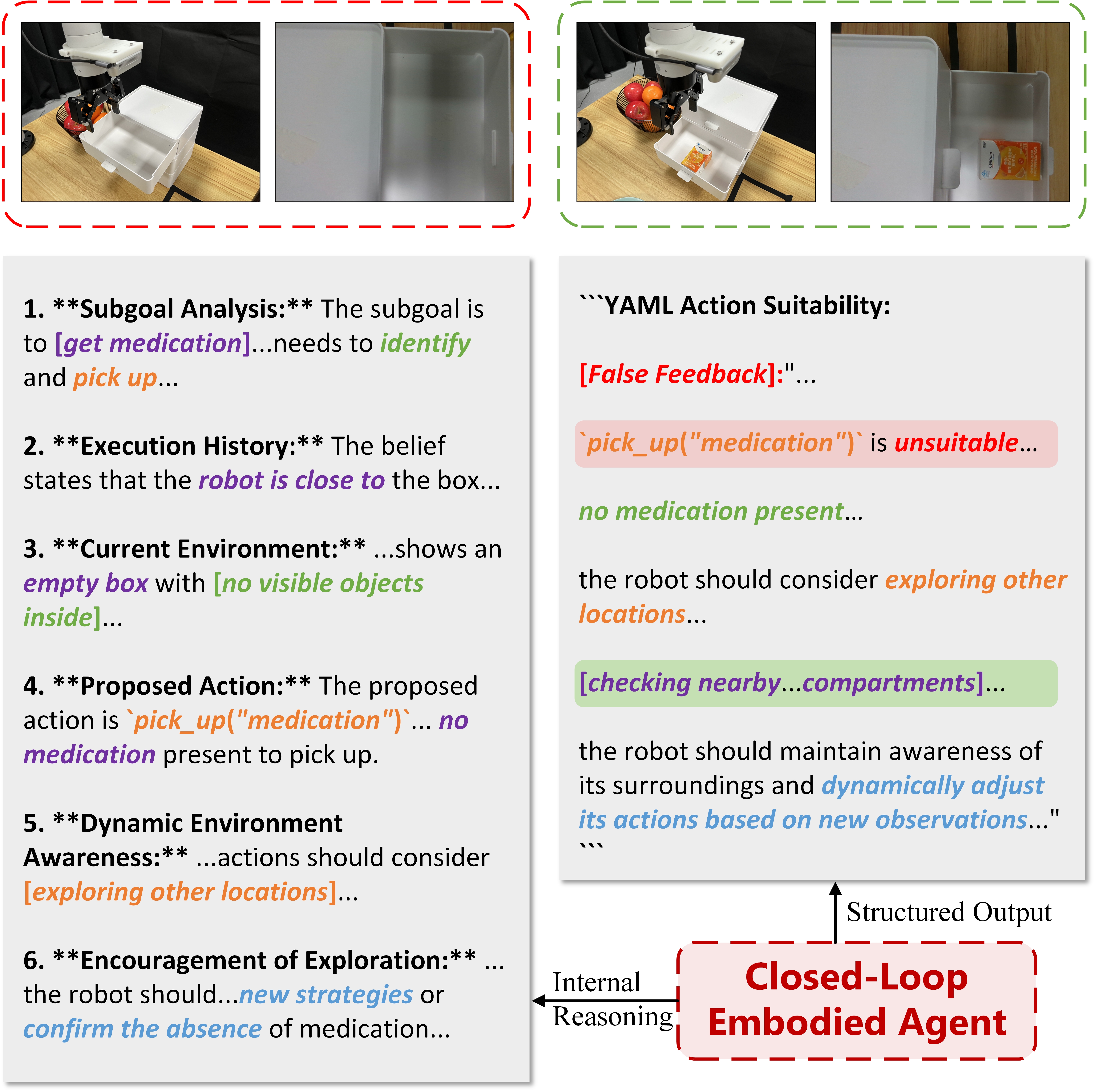}
   \vskip -0.05in
   \caption{\textbf{The reasoning and output of CLEA.} Unlike traditional failure-detection classification systems, CLEA performs internal reasoning upon receiving visual input and provides structured outputs. In the case where no medication is found in an empty drawer, the planner does not halt its intent. Instead, the critic suggests exploring alternative locations and provides the correct advice to check other compartments of the drawer, thereby guiding the successful completion of the task.
}
   \label{fig:pca}
\vskip -0.2in
\end{figure}
\section{Closed Loop Embodied Agent}
\label{sec:methodology}
\subsection{Overview of CLEA}

As illustrated in Fig. \ref{fig:onecol}, CLEA consists of three key components: an observer, a memory module, and a planner-critic agent. The observer bridges the image observations from the environment to the language model-based modules, converting visual input into textual descriptions. The memory module provides the agent with a belief of the current state, consisting of a history buffer to manage past interactions and a summarizer that generates beliefs based on this history. The planner-critic agent addresses dynamic planning in real-world embodied tasks. It is divided into two parts: the planner and the critic. The planner first sets a ``sub-goal" and plans a sequence of actions to achieve it, focusing on shorter action sequences rather than the entire task. During execution, the critic evaluates the ongoing plan at each step based on the belief state and the current observation. If the planned sequence is no longer suitable in the current situation, the planner will re-adjust the sub-goal and re-plan the action sequence accordingly.

The components of the CLEA framework work in unison to form a closed-loop planning model. This enables the system to handle dynamic and partially observable real-world environments by leveraging LLM's internal reasoning ability to analyze the robot's actions, detect discrepancies, and generate structured outputs. For example, in a task such as medication retrieval (as shown in Fig. \ref{fig:pca}), CLEA helps the robot understand its own behavior, recognize when its current strategy is suboptimal, and correct it in real-time.

\subsection{Observer}
\label{sec:obs}

% 由于开源 VLM 在多模态规划任务上有所不足，但是已有较强的对图片进行描述，总结能力。我们选择在 CLEA 的规划与记忆模块中使用没有多模态能力的 open-sourve LLM。故我们需要一个 VLM 将当前所见转述为文字。

%Due to the limitations of open-source VLMs in multimodal planning tasks, despite their strong capabilities in image description and summarization \cite{chiang2024chatbot}, we opt to use an open-source LLM without multimodal capabilities in the planning and memory module of CLEA. Therefore, we require a VLM to act as an observer, which can translate the current visual input into text. The observer aims to reconstruct the environment as accurately as possible based on image-based observation of the robot. For instance, the observer is tasked with selectively describing task-relevant objects and the relative relationships between these objects. The goal is to convert the image observation into a concise yet focused text-based observation. 
Open-source VLMs, while effective in image description and summarization \cite{chiang2024chatbot}, have inherent limitations in multimodal planning. As a result, CLEA employs an open-source LLM without multimodal capabilities for its planning and memory modules. To compensate for this, a dedicated VLM module serves as an observer, converting visual inputs into structured textual representations. The observer reconstructs the environment by selectively identifying task-relevant objects and capturing their spatial relationships, ensuring that the extracted information is both concise and semantically meaningful. We formally denote this text-based observation as \(o_i^t\). The observer module consists of a single VLM agent that processes an input image along with additional task-specific information and generates a structured textual output defined as: 
\begin{equation}
Obs: o_i \rightarrow o_i^t.
\end{equation}

\subsection{Memory}
\label{sec:memory} 

% 为了处理 POMDP 问题，我们选择使用 Memory 来帮助Agent在交互中获得对环境的 blief。与 observer 配合， Memory 模块得以以纯文字的形式记录robot与环境之间的交互。在实现上，Memory 基本可以看做一个 FIFO 的 Buffer，每个element就是 observation-action-feedback对。其中 observation 就是 observer 生成的 \(o_i^t\) , action 就是当前 Agent 调用的 API， feedback 就是 aciton 返回的关于机体内部“可能的”错误原因以及任务是否完成。

% History Buffer 是整个框架的基础，他manage了原始记忆数据，使得后序针对记忆数据进行总结等处理成为可能

%A first-in-first-out queue storing raw interaction tuples is defined as:
%\begin{equation}
%\mathcal{H} = \{h_i\}_{i=1}^n,
%\end{equation}
%where $h_i = (o_i, a_i)$ preserves the complete interaction history. This ensures no information loss while providing temporal grounding for downstream processing.

%We introduce a memory module to assist the agent in maintaining a belief about the environment during interactions. The memory module includes a history buffer and summarizer. 

We introduce a memory module to enable the agent to maintain a structured belief about the environment throughout its interactions. This module consists of two key components: a history buffer and a summarizer.

%The history buffer works with the observer to record the robot's interactions with the environment through textual logs. The history buffer can be viewed as a first-in-first-out buffer, where each element consists of an observation-action-feedback triple \((o_i^t, a_i, f_i)\). The observation \(o_i^t\) is generated by the observer, the action \(a_t\) is the action invoked by the agent in this iteration, and the feedback \(f_i\) represents the system's response, including potential internal errors and task completion status. The history buffer manages the raw memory data for downstream summarization. The memory buffer is formally defined as a queue containing interaction tuples:

The history buffer operates in conjunction with the observer to systematically record the robot's interactions with the environment in textual form. It functions as a first-in-first-out (FIFO) queue, where each entry is represented as an observation-action-feedback tuple \((o_i^t, a_i, f_i)\). The observation \(o_i^t\) is derived from the observer, the action \(a_i\) corresponds to the command executed by the agent at iteration \(i\), and the feedback \(f_i\) encapsulates the system's response, including internal errors and task completion status. This buffer serves as a repository for raw memory data, facilitating subsequent summarization. Formally, the history buffer is defined as:  
\begin{equation}
    H = \{h_i\}_{i=1}^n,
\end{equation}
where each element \( h_i = (o^t_i, a_i, f_i) \) records the interaction history between the robot and its environment at each iteration.

% 感觉以下statement没有任何意义，而且让人觉得奇怪
% The reconstruction not only ``summarizes" the history \(H\). But also ``recognizes" the environment. For example, the summarizer should summarize the spatial relationship of key objects.
The summarizer constructs a belief representation of the current environment based on the stored history \(H\). Leveraging the summarization capabilities of LLMs, the summarizer distills essential information from past interactions to infer the environment state \(s_i\) and the robot’s task progress. This information is consolidated into a structured textual representation, which is subsequently provided to the planning module. The summarization process transforms raw interaction histories into belief states, formally expressed as:  
\begin{equation}
    Sum: h_{[1:i-1]} \rightarrow b_i,
\end{equation}
where \(b_i\) represents the belief state, capturing compressed yet semantically rich historical information for CLEA. By maintaining a concise but informative belief representation, the summarizer facilitates efficient reasoning while preserving crucial temporal dependencies, thereby ensuring both memory efficiency and robust decision-making.  

%The summarizer generates belief about the current environment based on history \(H\). We utilize LLMs' summarize capability, asking an LLM to serve as the summarizer. The summarizer will try to reconstruct the state \(s_i\) of the environment and the robot-environment interaction. All the information will be integrated into a text as belief, which will be provided to the downstream planning module. The recognition process of the summarizer converts raw histories into belief states, which is given by:

%where $b_i$ represents the belief that compresses the historical information for CLEA. By maintaining compressed yet information-rich belief states, the summarizer enables efficient reasoning while retaining critical historical dependencies---striking a balance between memory efficiency and decision-making fidelity.

\subsection{Planner-Critic Agent}
\label{sec:planner-critic}

% planner-critic Agent 是本框架中 Planning Module。这个模块基于 Blief 与给定的其他环境信息，实现了一种闭环规划机制。该机制由 planner 的分层的规划机制与 critic 的动态监督机制组成。首先，与当前部分框架不同（cite VLM-TAMP）,为了适应真实环境中的动态性，我们提出了一种分层的规划机制。我们并不会将 task 在最初就分解为数个sub-goal，而是在 planner 每次规划之前先决定 ``下一步需要完成的 sub-goal"。并以该 ``sub-goal" 为目标进行 action sequence 规划。故我们将 task 分解为两个层级。 ``sub-goal" 与 ``action sequence"。对于 ``sub-goal" 层，我们是step-wise的规划，即每次规划仅维护一个 ``sub-goal".这个好处是下一个 ``sub-goal" 是可以基于环境变化动态调整的。而 ``action sequence" 会一步输出完成的规划 sequence 并顺序执行。这个设定的好处是由于每次仅需以达成sub-goal为目标进行规划，故规划步数较短但依然能保证规划的连续性。这缓解了 LLM 规划能力的不足（cite LLM can not plan）。第二，我们提出了动态监督机制。在每一步执行之前，我们都会要求 critic 模块基于最新的 belief 与 observation 判断继续按照原计划执行是否合理。是否出现了意外情况等。若判断现行计划已经不适合最新环境则要求 planner 重新制定 ``Sub-goal" 与 action sequence. The planner-critic module integrates memory with systematic reasoning to generate robust action plans. As shown in Figure \ref{fig:onecol}, it operates via two cooperative components:

The planner-critic agent module is responsible for closed-loop decision-making and consists of two key components: the planner and the critic.  

The planner module employs a hierarchical planning mechanism, leveraging the belief state and available environmental information to generate a new sub-goal along with the corresponding action sequence. The planner is implemented as a single LLM agent utilizing Chain-of-Thought (CoT) prompting \cite{wei2022chain} to infer sub-goals and associated action sequences. By providing structured prompts, the LLM receives relevant contextual information and subsequently outputs the current sub-goal along with a sequence of executable actions. These actions are selected from a predefined skill pool, as illustrated in Fig.~\ref{fig:onecol}, enabling the CLEA agent to interact with the robotic platform in real-world scenarios. Formally, at time step \(i\), the planner generates a sub-goal \(g_i\) and an action sequence \(\{\hat{a}_{i:k}\}_{k=1}^m\) based on the belief state \(b_i\), which can be expressed as:  
\begin{equation}
    P(b_i, o_i) \rightarrow (g_i, \{\hat{a}_{i:k}\}_{k=i}^m).
\end{equation}
Each action \(a_i\) within the sequence corresponds to a predefined function call from the skill pool, which consists of modular Python functions. For each action, the planner invokes the appropriate skill function with relevant parameters, ensuring structured execution.  

The critic module incorporates a VLM agent with CoT prompting to assess the feasibility of the proposed actions in real time. The critic determines whether the planned action at each iteration is appropriate given the current state of the environment. Through empirical analysis, we observe that providing the critic with the original image, rather than a text-based observation, is crucial for accurate evaluation. Since the critic must infer spatial relationships, object states, and potential occlusions from fine-grained visual details, relying solely on abstract textual descriptions proves insufficient.  

Prior to the execution of each action, the critic evaluates its validity by considering the belief state and the current visual observation. If the action is deemed feasible, the agent proceeds with execution. Conversely, if the action is found to be unsuitable, the critic generates corrective feedback, prompting the planner to refine the sub-goal and recompute the action sequence. This iterative validation process is formally defined as follows:  
\begin{equation}
    C(\hat{a}_{i}, b_i, o_i) \rightarrow (p_i, f_i),
\end{equation}
where \(p_i\) is a binary validity flag (\(p_i \in \{\text{true}, \text{false}\}\)), and \(f_i\) represents the feedback signal. If \(p_i = \text{false}\), the planner utilizes \(f_i\) along with the updated belief state \(b_i\) to regenerate an improved action sequence \(\{\hat{a}_{j:k}\}\), thereby forming an adaptive refinement loop that dynamically adjusts the plan in response to environmental variations.  

By integrating memory-driven reasoning with real-time environmental feedback, the planner-critic framework facilitates adaptive task execution, enhancing the robustness and flexibility of the planning process. The synergy between belief states, visual observations, and iterative planning ensures that the agent can effectively navigate dynamic environments while adhering to the constraints of real-world robotic systems.

\section{Experiments}

% Visualization of the experimental environment.
% 字大一点, 至少和caption一样大
% 不要 such as

\begin{figure}[htbp]
\centering
   \includegraphics[width=0.99\linewidth]{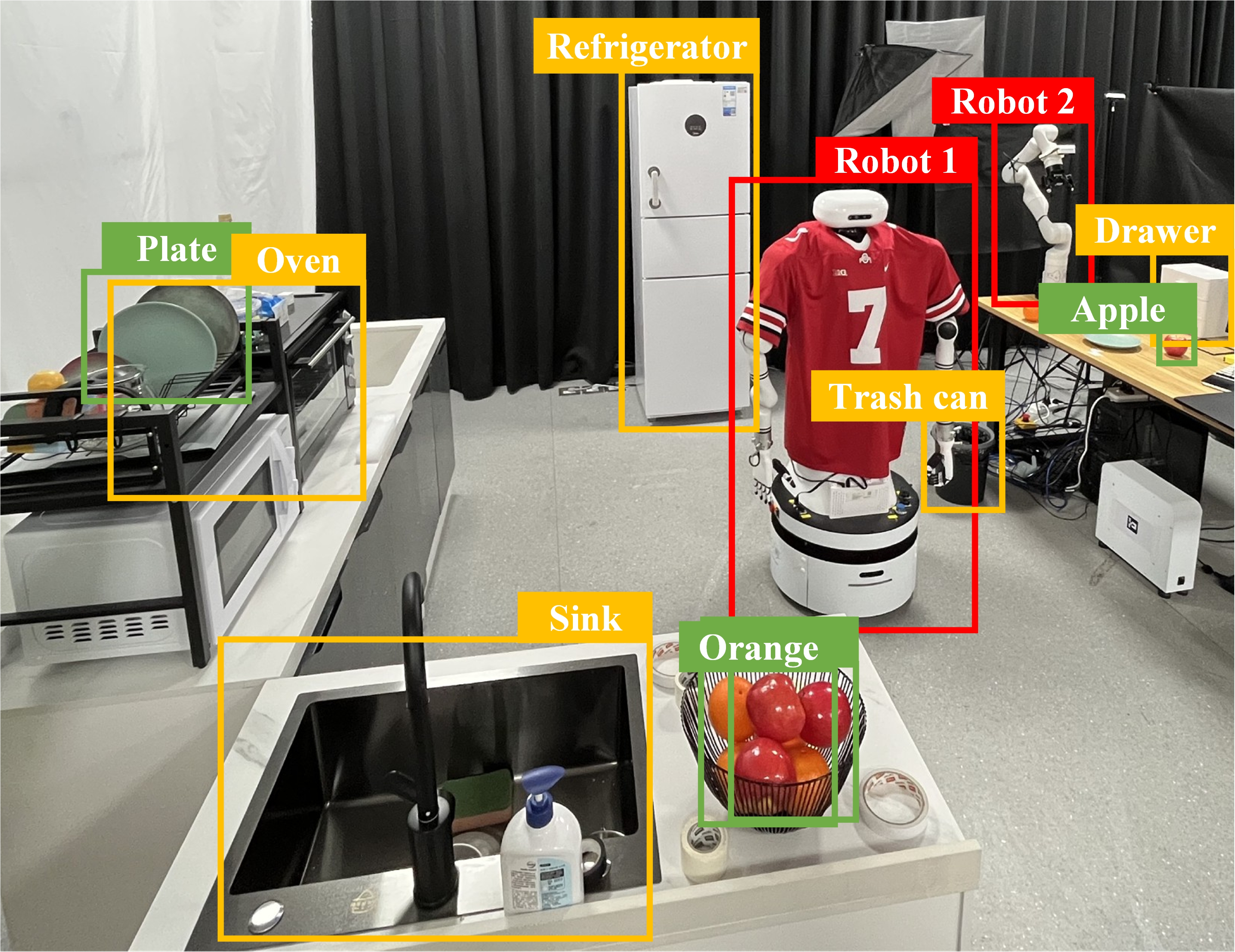}
   \vskip -0.05in
   \caption{Visualization of the experimental environment.}
   \label{fig:env}
\vskip -0.2in
\end{figure}

\subsection{Experimental Settings}

% 我们在真实环境中部署了两个机器人。我们要求 CLEA 同时控制两个机器人，并在使用时接受两个机器人的动作反馈与Observation。两个机器人分别是一个双臂机器人与一个机械臂。双臂机器人能移动与操作物品，而单臂机器人位置固定，但是能覆盖双臂机器人难以达到的死角。通过测试 CLEA 利用两个机器人的不同性质，通过动态的action 规划解决问题的能力，我们能展示 CLEA 在面对动态且部分可观测的环境中进行长期闭环规划的能力。

% 我们的测试环境是一个小型开放式厨房环境。我们的环境包括 10 个可操作物品，四个不可见空间（如冰箱，抽屉等）与三种操作。我们为两个不同的机器人分别定义了不同的动作function，具体请见 （图表）。为了充分测试 CLEA 的长期闭环规划能力，我们设置了三种不同的任务：(1) 寻找任务。我们将要求 CLEA 规划两个机器人在环境中配合寻找两个物品并放入指定位置。（2）操作任务。我们将提供两个物品所在的位置（可能并不可见），要求 CLEA 对两个物品进行操作。（3）组合任务。我们将结合寻找与操作任务。我们要求机器人在环境中寻找两个物品并对两个物品进行操作。对于每种任务，我们都运行三个 trial，每个 trial 中都要求其寻找不同物品以测试其规划的鲁棒性。通过以上三种任务，我们将测试 CLEA 在完成厨房场景下长程闭环规划能力。

% 我们选择三个 Criteria：(1) SR 成功率, (2) AS 平均得分率与 (3) 成功步数。 

{\bf Environment description.} In this study, we deploy two robots in a real-world environment. We require CLEA to simultaneously control both robots, receiving feedback and image observations from each robot's actions. The two robots consist of a dual-arm robot and a single-arm robot. The dual-arm robot is capable of both movement and manipulation, while the single-arm robot remains stationary but covers areas that are difficult for the dual-arm robot to reach. By testing CLEA's ability to leverage the distinct characteristics of both robots and dynamically plan actions, we demonstrate CLEA's capacity for long-term closed-loop planning in a dynamic and partially observable environment. Our test environment is a small, open kitchen setup. This environment contains 10 manipulable objects, 4 closed containers (e.g., a refrigerator, two drawers and an oven), two open spaces (a table and a sink), and 3 interactive devices (an oven, a refrigerator and a garbage can). The overall environment is shown in Fig. \ref{fig:env}.

{\bf Task definition.} To thoroughly evaluate CLEA's long-term planning ability, we define three types of tasks: (1) Search task: CLEA coordinates the two robots in a collaborative search for two distinct objects. Direct visualization of objects is confirmed as ``find". (2) Manipulation task: Given the locations of two objects (which may be partially occluded), CLEA is required to manipulate both objects. (3) Integration task: This task integrates the search and manipulation tasks. CLEA must find and manipulate one or two distinct objects within the environment.

{\bf Evaluation metric.} 
%For each task, we run three trials with different environment settings, allowing us to test the robustness of the planning system. Through these three tasks, we assess CLEA's ability to perform long-term closed-loop planning in a kitchen environment. We select three evaluation criteria: (1) Success Rate (SR): The proportion of successful task completions. (2) Average Score (AS): The average performance score across all trials. We count each correct action as one point.
We conduct three trials for each task under varying environmental conditions to evaluate the robustness of the planning system. These trials enable a comprehensive assessment of CLEA’s capability for long-term closed-loop planning within a kitchen environment. To quantitatively measure performance, we define two evaluation metrics: (1) Success Rate (SR), which represents the proportion of successfully completed tasks; (2) Average Score (AS), calculated as the mean performance score across all trials, where each correctly executed action contributes one point.

{\bf Baseline agent.}
% CLEA w/o critic: critic 是我们模块中重要部分，通过对 critic 进行 ablation。虽然此时在每个 subgoal 对应的 aciton list 执行完后 planner 依然有重新规划的机会，但是在 aciton list 执行期间，agent 并没有机会针对突发情况进行规划。
% Open-loop planner：Open-loop planner 为单个 VLM，该 planner 将会需要规划完整的 Action List，并没有 CLEA 的闭环规划与subgoal机制。该 Baseline 代表了先前具身规划框架实现 \cite{...}
% baseline：我们 
% We consider an open-loop planning agent for performance comparison, which generates an action sequence through a VLM with additional refinement. When the action sequence is fully executed or an inaccurate action is encountered, the VLM will re-plan its actions (one re-plan opportunity). This reveals how CLEA overcomes the brittleness in traditional static planning approaches. %We give VLM three re-try opportunities.
To benchmark performance, we implement an open-loop planning agent that generates action sequences using a VLM with additional refinement \cite{yang2024guiding}. The agent follows a predefined plan until either the sequence is fully executed or an erroneous action is encountered. In the latter case, the VLM is granted a single opportunity to re-plan its actions. This baseline highlights the limitations of static planning approaches and demonstrates CLEA’s ability to enhance robustness through closed-loop adaptation.

{\bf Ablation studies.}
%We dissect our technical contributions by examining a degraded variant, CLEA w/o critic, where the critic module is removed. While this version retains sub-goal generation capabilities, its inability to detect execution errors or recover from unexpected failures—such as object misplacement or environmental changes—conclusively demonstrates the critic's critical role in enabling real-time plan refinement. These evaluations systematically validate both CLEA's superiority over existing methods and the necessity of its core components.
To systematically assess the impact of our proposed framework, we conduct an ablation study by evaluating a degraded variant, CLEA w/o critic, in which the critic module is removed. While this variant maintains the ability to generate sub-goals, it lacks the capability to detect execution errors or adapt to unexpected failures, such as object misplacement or environmental changes. The inability to perform real-time plan refinement underscores the critical role of the critic module in ensuring adaptive task execution. These evaluations provide empirical validation of both CLEA’s superiority over existing methods and the necessity of its core components.

\begin{table}[htbp]
\centering
\caption{Predefined skill pool in the environment}
\label{tab:skill}
\begin{tabular}{|l|l|}
\hline
\textbf{Skill pool} & \textbf{Description} \\ \hline
open(robot, openable\_object) &  robot  open  object  \\ \hline
close(robot, openable\_object) &  robot  close  object  \\ \hline
pick\_from(robot, object, space) &  robot  pick  object  from  space  \\ \hline
release\_to(robot, space) &  robot  release the object on its hand \\
 & to  space  \\ \hline
go\_to(robot, navi\_point) &  robot  navigate to  navigation point  \\ \hline
\end{tabular}
\end{table}

% latex中表格编号和ref引用编号
% 我们使用开源大模型 power CLEA 的各个模块。对于 LLM 模块（即 Summarizer 与 planner 模块），我们使用 Qwen2.5-72b-instruct 模型，对于 VLM 模块，即 observer 与 critic 模块，我们使用 Qwen2.5-72b-vl-instruct。 两个模型在各项测试上都有接近先进闭源模型的能力。所有模型都使用阿里云百炼 API，temperature 统一设置为 0.2。CLEA 通过 http 协议与机器人真机进行交互。
{\bf Implementation details.}
Our framework leverages open-source LLM and VLM models to power its core modules. Specifically, the Qwen2.5-72B-Instruct model \cite{yang2024guiding} is employed for LLM-based components, including the summarizer and planner, while the Qwen2.5-72B-VL-Instruct model \cite{bai2025qwen2} is used for VLM-driven modules, such as the observer and critic. The action space is defined by a predefined skill pool, as detailed in Table \ref{tab:skill}. The experimental hardware setup consists of a Kinova Gen3 7-DoF fixed-base manipulator, an RM65-B dual-arm mobile manipulator, and multiple depth cameras. For motion planning, we employ the RRT-Connect algorithm, implemented within the MoveIt framework \cite{Coleman2014Reducing}, across all robotic platforms. The robot's perception system integrates the YOLOv11 segmentation model \cite{khanam2024yolov11} alongside particle filter-based localization and mapping techniques, ensuring robust scene understanding and navigation.

% We utilize open-source LLM/VLM models. Specifically, the Qwen2.5-72b-instruct model \cite{yang2024guiding} powers the LLM modules, including the summarizer and planner, while the Qwen2.5-72b-vl-instruct model \cite{bai2025qwen2} drives the VLM modules, such as the observer and critic. The action space is defined by a predefined skill pool, as shown in Table \ref{tab:skill}. The experimental hardware setup includes a Kinova Gen3 7-DoF fixed-base manipulator, an RM65-B dual-arm mobile manipulator, and multiple depth cameras. For motion planning, the RRT-connect algorithm, implemented in MoveIt \cite{Coleman2014Reducing}, is applied across all robotic platforms. The robot's perception system leverages the YOLOv11 \cite{khanam2024yolov11} segmentation model and particle filter-based localization and mapping techniques.

\subsection{Comparison Results}

\begin{figure}[htbp]
\centering
   \includegraphics[width=0.99\linewidth]{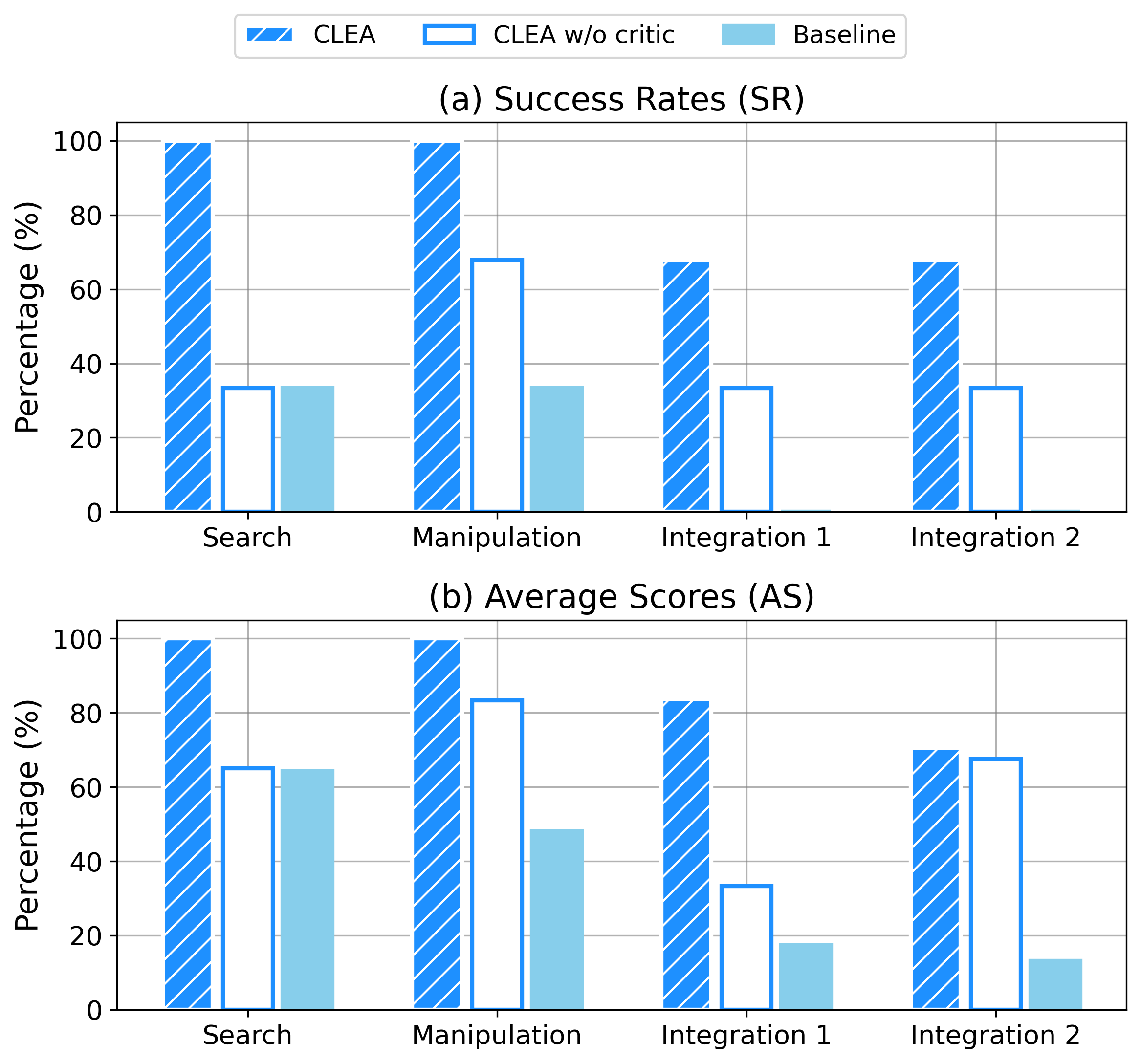}
   \vskip -0.05in
   \caption{Comparisons among the CLEA, the ablation, and the baseline agent.}
   \label{fig:baseline}
\vskip -0.1in
\end{figure}

\begin{table}[htbp]
\caption{Reasons and ratios when critic raise \(p_i = \text{false}\)}
\centering
\scriptsize
\resizebox{1.0\linewidth}{!}{
    \begin{tabular}{cccc}
        \toprule
             Type & Reason & Count & Total \\ 
        \midrule
        \multirow{4}{*}{\textbf{Critic}}
                & Outdated actions
                    & 8 &  44.4\%  \\
                & Redundant actions
                    & 6 &  33.3\% \\
                & Invalid actions
                    & 2 &  11.1\%  \\
                & Wrong planning
                    & 2 &  11.1\%  \\
        \bottomrule
    \end{tabular}

}

\label{tab:critic_reasons}
\end{table}

We conduct a total of 12 real-world experiments. Based on the data from (a) SR and (b) AS. Our experimental results are shown in Fig. \ref{fig:baseline}. The proposed CLEA framework demonstrated significant improvements over baseline methods. On average, CLEA outperformed the baseline by an increase of 67.3\% in SR and 53.8\% in AC. Additionally, CLEA was also more effective than CLEA w/o critic, showing a marked improvement of 42.0\% in SR and 26.3\% in AC. These results underscore the effectiveness of the CLEA framework.

Comparing CLEA with the baseline, we observe that while the baseline agent can accomplish basic searching and manipulation tasks, it consistently fails in the integration 1 \& 2 tasks. In contrast, CLEA successfully completes the majority of tasks, demonstrating the effectiveness of closed-loop planning in dynamic real-world environments.

In the search task, the baseline agent often fails to comprehensively explore potential locations where target objects may be located. This limitation likely arises from the absence of a memory mechanism, which prevents the agent from reconstructing past experiences. Memory is particularly crucial for search tasks where target object locations remain uncertain. In the manipulation task, we find that the open-loop long-term planning capabilities of the baseline LLM are insufficient, occasionally leading to logical inconsistencies in the planned action sequence. In the integration task, which requires both searching and manipulation, the baseline’s inability to effectively perform both subtasks results in complete task failure.

For CLEA without the critic module, the absence of dynamic action evaluation leads to the execution of redundant operations, some of which directly result in task failure. In long-horizon searching and manipulation tasks, the lack of real-time correction from the critic causes CLEA to frequently encounter issues such as attempting to grasp non-existent objects or interacting with unopened containers (e.g., trying to retrieve an item from a closed refrigerator). These errors significantly reduce both the success rate and the overall task performance score, highlighting the necessity of the critic module in refining action sequences and ensuring task feasibility.

As shown in Table \ref{tab:critic_reasons}, we examine several common errors detected by the critic module:

\textit{1) Outdated actions: }The planner may make assumptions about the environment that later prove to be incorrect. For instance, the planner might assume that an object is located in a specific place, such as inside the refrigerator. If this assumption turns out to be false, the subsequent plan becomes outdated. This is a common issue in dynamic environments, and the critic plays a vital role in detecting such outdated actions, which highlights its importance in improving CLEA's performance in ever-changing contexts.

\textit{2) Redundant actions:} Occasionally, the planner overlooks information provided by the belief state, leading to planning actions that have already been executed or are no longer necessary. The critic can identify these redundant actions, preventing unnecessary task execution and improving overall efficiency.

\textit{3) Invalid actions:} CLEA may output actions that do not conform to the required API format. Table \ref{tab:critic_reasons} reveals that the critic detected two invalid actions. This suggests that the critic is able to detect invalid actions that remain imperfect.

\textit{4) Incorrect planning:} Due to limitations in the LLM planner, some generated plans may be unfeasible. The critic can detect these planning errors and, in some cases, provide suggestions for creating more practical plans, helping to guide the planning process toward more effective solutions.

These findings emphasize the critical importance of closed-loop planning with dynamic evaluation in robotic systems operating in complex, real-world environments. The integration of a dynamic feedback loop between the planner and the critic proves essential for the robustness and adaptability of CLEA in such settings.

\subsection{Failure Analysis}
\begin{table}[htbp]
\caption{Reasons and ratios of failure caused by CLEA}
\centering
\scriptsize
\resizebox{1.0\linewidth}{!}{
    \begin{tabular}{cccc}
        \toprule
             Type & Reason & Count & Total \\ 
        \midrule
            \multirow{3}{*}{\textbf{Failure}}
                & Invalid actions 
                    & 9 &  45\% \\
                & Critic failures
                    & 8 &  26.7\%  \\
                & Multi-robot collaboration 
                    & 3 &  15\%  \\
        \bottomrule
    \end{tabular}

}

\label{tab:failure_reasons}
\end{table}

As shown in Table \ref{tab:failure_reasons}, we identify and analyze three common failure modes within the CLEA framework. While these anomalies may not directly cause task failure, they can hinder overall task completion.

\textit{1) Action API limitations in real robots:} The most frequent failure occurs due to invalid actions. These arise when the LLM fails to generate actions in the correct Python function call format. Although invalid actions do not immediately lead to task failure, they reduce the efficiency of CLEA's performance. This issue likely stems from the limitations of the predefined action format, which restricts the planner's flexibility and limits its ability to generate optimal action sequences.

\textit{2) Critic failures:} While the critic is generally effective in detecting execution errors, there are instances where it fails to identify improper actions. For example, the critic may erroneously halt a valid plan or fail to recognize an error in the action sequence. This limitation is likely due to the constraints in the VLM's perceptual capabilities, which may not always capture the full range of environmental nuances required for accurate error detection.

\textit{3) Multi-robot collaboration issues:} Certain tasks necessitate coordination between two robots. In these scenarios, we observe that CLEA occasionally struggles to correctly interpret and manage inter-robot collaboration. For example, CLEA may assign actions intended for one robot to another. This issue may arise because LLMs are not particularly adept at understanding and managing complex inter-robot relationships.

\section{Conclusion and Future Work}

We introduce CLEA, a closed-loop framework designed for strategy adjustment in embodied systems, and evaluate it through search, manipulation, and integration tasks in a real-world kitchen environment. In comparison to open-loop frameworks, CLEA demonstrates significant improvements in both task completion and error recovery for long-horizon tasks. Looking ahead, we aim to integrate more advanced reasoning models to enhance the planning capabilities of the system while also experimenting with smaller models to improve the efficiency of LLM/VLM. Furthermore, we intend to incorporate Visual Language Action (VLA) models to bolster the robot's operational robustness during task execution. Currently, the limited action space poses challenges; even when errors and their causes are identified, recovering from failures at the action level remains difficult without function calls. To address this, we plan to leverage VLA technology at the lower levels, providing a richer and more flexible action space for improved error recovery. 

                                  % the height of the last page by a suitable amount.

%%%%%%%%%%%%%%%%%%%%%%%%%%%%%%%%%%%%%%%%%%%%%%%%%%%%%%%%%%%%%%%%%%%%%%%%%%%%%%%%

%%%%%%%%%%%%%%%%%%%%%%%%%%%%%%%%%%%%%%%%%%%%%%%%%%%%%%%%%%%%%%%%%%%%%%%%%%%%%%%%

\bibliographystyle{IEEEtran}
\bibliography{references}

\end{document}